\documentclass{article}
\usepackage{algorithm}
\usepackage{algorithmic}

\usepackage{microtype}
\usepackage{graphicx}
\usepackage{subfigure}
\usepackage{booktabs} 

\usepackage{hyperref}

\usepackage[accepted]{icml2024}

\usepackage{amsmath}
\usepackage{amssymb}
\usepackage{mathtools}
\usepackage{amsthm}

\usepackage[capitalize,noabbrev]{cleveref}
\usepackage{amsmath}
\usepackage{enumitem}
\usepackage{multirow}
\usepackage{multicol}
\usepackage{ulem}
\definecolor{Gray}{gray}{0.9}

\theoremstyle{plain}

\theoremstyle{definition}

\theoremstyle{remark}

\usepackage[textsize=tiny]{todonotes}

\icmltitlerunning{Alignment is All You Need: A Training-free Augmentation Strategy for Pose-guided Video Generation}

\begin{document}

\twocolumn[
\icmltitle{Alignment is All You Need: A Training-free Augmentation Strategy for Pose-guided Video Generation}



\icmlsetsymbol{equal}{*}

\begin{icmlauthorlist}
\icmlauthor{Xiaoyu Jin}{equal,yyy,xxx}
\icmlauthor{Zunnan Xu}{equal,yyy,xxx}
\icmlauthor{Mingwen Ou}{yyy,xxx}
\icmlauthor{Wenming Yang$^{\dag}$}{yyy}
\end{icmlauthorlist}

\icmlaffiliation{yyy}{Shenzhen International Graduate School, Tsinghua University}
\icmlaffiliation{xxx}{Work done during internship at Tencent}

\icmlcorrespondingauthor{Wenming Yang}{yang.wenming@sz.tsinghua.edu}

\icmlkeywords{Machine Learning, ICML}

\vskip 0.3in
]

\printAffiliationsAndNotice{\icmlEqualContribution} 

\begin{abstract}
Character animation is a transformative field in computer graphics and vision, enabling dynamic and realistic video animations from static images. Despite advancements, maintaining appearance consistency in animations remains a challenge. Our approach addresses this by introducing a training-free framework that ensures the generated video sequence preserves the reference image's subtleties, such as physique and proportions, through a dual alignment strategy. We decouple skeletal and motion priors from pose information, enabling precise control over animation generation. Our method also improves pixel-level alignment for conditional control from the reference character, enhancing the temporal consistency and visual cohesion of animations. Our method significantly enhances the quality of video generation without the need for large datasets or expensive computational resources.
\end{abstract}

\section{Introduction}
\label{intro}
Character animation is a task in the fields of computer graphics and computer vision to enable the shift from static images to dynamic, realistic video animations. This technology has significant implications for various industries such as entertainment, social media, virtual reality, and other immersive digital experiences, providing more engaging and customized visual experiences. A key challenge in this area is maintaining appearance consistency and fidelity in animated sequences, as these aspects are essential for the realism and overall quality of the produced content. 

Previous endeavors in character animation have advanced the transformation of static images into dynamic content. Traditional graphic techniques have been enhanced by data-driven models leveraging extensive visual datasets for more cost-effective solutions. While GAN-based methods~\cite{gan,wgan,stylegan} show potential in creating realistic details, they face challenges with motion transfer and maintaining subject identity across poses. Conversely, diffusion-based models~\cite{denoising,dreampose,ldm}, while capable of producing visually plausible animations, are susceptible to appearance inconsistencies, resulting in  unnatural limb proportions and sub-optimal effects when there are significant differences between the reference image physique and the pose used for generation~\cite{animatediff,xu2023magicanimate,li2023g2l,hu2023animate}.
Our approach introduces a training-free framework that prioritizes appearance consistency. Unlike existing methods that ignore appearance details during animation, our method ensures the generated video sequence stays true to the motion while preserving subtleties of the reference image, like the subject's physique—accurately reflecting their height, build, and proportions in the image. Through a dual alignment strategy, our method can create animations that show both appearance consistency and fidelity to the reference image.

Building on the insights from previous research, our method introduces a dual alignment strategy that re-envisions the relationship between reference images and pose data. A core element of our innovation lies in the separation of skeletal and motion priors from the pose information itself. We identify the essential cues present in the key points representations, such as skeletal position, length and angular variances, which reflect an individual's body information and motion tendencies.
By utilizing efficient linear matrix operations, our approach distinguishes the identity information and the movement information of the skeletal sequences. This enables the transfer of skeletal data from a reference image to the driving pose sequences while preserving the intrinsic motion characteristics of poses. This allows for precise control over the generation, ensuring that the animation faithfully reproduces the physique of reference character while maintaining a resemblance to the motions of the pose sequences.
Furthermore, acknowledging the importance of accurate pixel-level alignment for conditional control, we improve the reference image to kickstart an animation that closely aligns with the initial frame of the driving pose video. This enhancement utilizes the information stored in current diffusion models to direct the reference image to mimic the motion of the starting pose. The outcome is an improved alignment between the reference image and the driving pose video, establishing the foundation for a temporally consistent and visually cohesive animation sequence.
Our main contributions are as follows:
\begin{itemize}[leftmargin=*,noitemsep,nolistsep]
    \item We introduce a training-free augmentation strategy for pose-guided animation generation that avoid the need for using large video datasets and expensive GPU resources.
    \item We propose a novel dual alignment method that can be seamlessly integrated into pose-guided generative models to enhance the quality of generated videos.
    \item Experiments demonstrate that our method can effectively enhance the quality of character animation generation.
\end{itemize}

\section{Related Work}
\label{related}
\subsection{Diffusion Model for Image Generation}
In the domain of text-to-image synthesis, diffusion-based models have indeed set new benchmarks for generation quality and have become a central focus of research. These models, such as DALL-E 2~\cite{dalle2}, Imagen~\cite{imagen}, Latent Diffusion Model (LDM)~\cite{ldm}, Glide~\cite{glide}, eDiffi~\cite{ediffi}, and Composer~\cite{composer}, have demonstrated the ability to produce high-quality and diverse image outputs from textual descriptions~\cite{lu2024coarse,li2024exploiting}.
The Latent Diffusion Model (LDM) has introduced a method of denoising in the latent space to reduce computational complexity while maintaining the quality of generated images. This offers an effective and efficient approach to image synthesis.
Further advancements have been made in controlling the visual generation process. 
With the development of parameter-efficient tuning methods~\cite{xu2023bridging,xu2024enhancing}, models like ControlNet~\cite{controlnet} and T2I-Adapter~\cite{t2iadapter} have integrated additional encoding layers to enhance the controllability of the generation process. This enhancement allows for conditional generation based on various factors such as pose, mask, edge, and depth information.
Building upon these capabilities, some studies have explored image generation that is conditioned on specific image-related inputs. For instance, IP-Adapter~\cite{ip} enables diffusion models to generate images that incorporate content specified by an image prompt. ObjectStitch~\cite{objectstitch} and Paint-by-Example~\cite{paint} utilize the capabilities of CLIP to propose diffusion-based methods for image editing under given image conditions.
In the context of fashion and virtual try-on applications, TryonDiffusion~\cite{tryondiffusion} applies diffusion models to the task of virtual apparel try-on and introduces an innovative Parallel-UNet structure to enhance the process.
These developments highlight the rapid progress and innovation in the field of text-to-image generation. Diffusion-based methods are not only pushing the boundaries of what is possible but also expanding the horizons of controllability and applicability in diverse scenarios.

\subsection{Pose Guidance in Character Animation Generation}
The success of diffusion models in text-to-image synthesis has significantly influenced text-to-video research~\cite{text2videozero,fatezero,cogvideo,tuneavideo,rerender,xu2024chain,gen1,makeavideo,vdm}, particularly in model structure and the incorporation of pose guidance. In the context of pose guidance, DWpose~\cite{dwpose} offers an enhanced alternative to OpenPose~\cite{openpose}, providing more accurate and expressive skeletons that are beneficial for high-quality image generation.  DensePose~\cite{densepose} establishes dense correspondences between images and surface representations, which is crucial for detailed pose guidance. The SMPL model~\cite{SMPL:2015}, known for its realistic human representation, is widely used for pose and shape analysis in character animation~\cite{li2023jotr,yang2024freetalker,xu2024mambatalk}. It serves as essential ground truth for neural networks and is considered in our approach for reconstructing poses and shapes, providing a comprehensive foundation for appearance alignment and pose guidance in video generation.
Our approach draws inspiration from these methods by focusing on decoupling the identity and the movement from DWpose/Openpose guidance in the video generation pipeline. This ensures that the generated animations maintain coherence in appearance with the reference image.

\section{Methodology}
\label{method}
To create pose-guided personalized videos in a training-free setting, we introduce a simple yet effective framework in Section~\ref{method}. In Section~\ref{subsec:Framework}, we describe the settings and architecture of our pipeline. Section~\ref{subsec:Pose Adapter} details our training-free Skeleton based Pose Adapter method. Finally, in Section~\ref{subsec:Kickstart Alignment}, we present our Kickstart Alignment Strategy.
\begin{figure*}[ht]
	\centering
	\includegraphics[width = 1.0\linewidth]{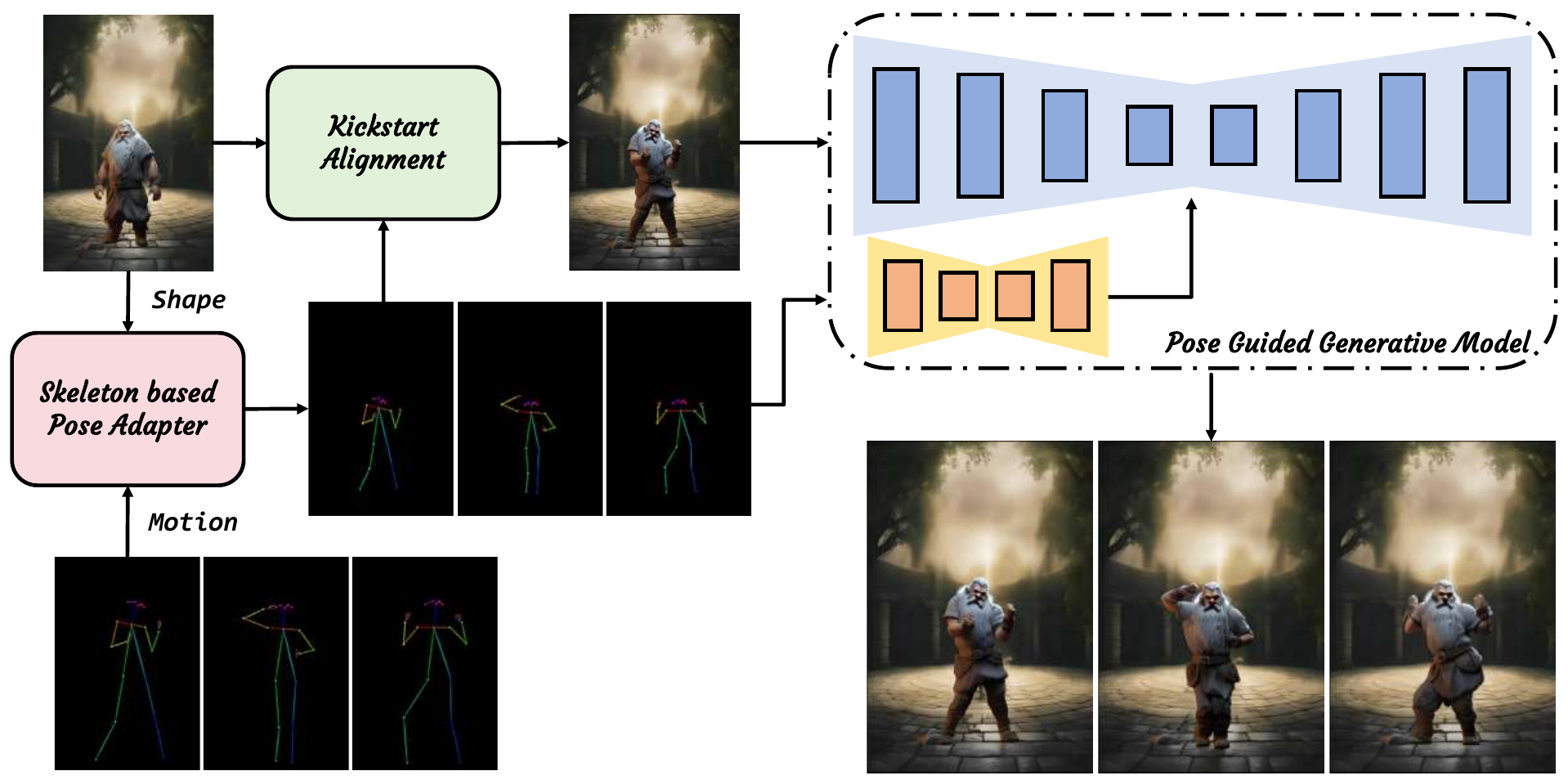}
	\caption{The overall framework of our method. 
 The architecture of our method is composed of two key components: Skeleton based Pose Adapter and Kickstart Alignment. These components work together to refine the pose used for driving video generation and the reference image. These refined control conditions are then inputted into the existing pose-guided video generation model, enabling dynamic and realistic video animations with increased consistency and fidelity.
}
\label{fig:framework}
\end{figure*}

\subsection{Settings and Framework}
\label{subsec:Framework}
Given a reference character image $I_r$ and a template pose sequence $\mathbf{P} = \{\mathbf{p}_a \}_{a=1}^M$ consisting of $M$ frames, our goal is to generate a high-quality video $\mathbf{V} = \{\mathbf{v}_a \}_{a=1}^M$. This video should be of superior fidelity, exhibit exceptional faithfulness to $I_r$, and accurately align with $\mathbf{P}$.

As illustrated in Figure~\ref{fig:framework}, our framework builds upon the existing pose-guided video generation model. Our contribution is entirely training-free, manipulating the input reference image and pose sequence without participating in the training of the main network. Specifically, the Skeleton based Pose Adapter decouples and embeds the identity information from the input pose sequence. The aligned pose is then used to transfer the input image into another one with the gesture of the initial pose, while preserving the identity information. Similar to other diffusion-based video generation methods, the U-net employs multiple frames of noise, along with a reference image and pose sequence with detailed identity information, to generate vivid personalized videos.

\begin{algorithm*}[t]
	\caption{PoseAdapter}
	\label{alg:Pose Adapter}
	\textbf{Input}: Template pose sequence: $\mathbf{poses1}$,  reference image pose: $pose2$. \\
	\textbf{Output}: Transformed pose image $\mathbf{kps\_results2}$, transformed pose array $\mathbf{poses2}$.
	\begin{algorithmic}[1]
        \STATE Scale $pose2$ and $\mathbf{poses1}$ by their image shape to match the real size.
        \STATE Calculate $edge\_ratios$ using the coordinates and limbSeq of $\mathbf{poses1}$ and $pose2$.
        \STATE Init $\mathbf{pose2}$ and $\mathbf{kps\_results2}$.
        \FOR{$pose1$ in $\mathbf{poses1}$}
            \STATE Update body positions of $pose2$ with $edge\_ratios$ and $pose1$.
            \STATE Update hand positions of $pose2$ with $pose1$, updated body positions and $edge\_ratios$.
            \STATE Normalize $pose2$ to draw the pose on a canvas to get transformed pose image2.
            \STATE Add pose image2 to $\mathbf{kps\_results}$ and add $pose2$ to $\mathbf{pose2}$.
        \ENDFOR
        \STATE Output $\mathbf{kps\_results2}$, $\mathbf{pose2}$
	\end{algorithmic}
\end{algorithm*}

\subsection{Skeleton based Pose Adapter}
\label{subsec:Pose Adapter}
In pose-guided video generation tasks, a skeleton-based human pose estimation model~\cite{dwpose,openpose} is usually employed to extract the pose sequence $\mathbf{P}$ from a template video. This sequence combines action sequence information with the identity information, such as the physique, position, and distance from the camera.

However, the identity information embedded within the template video is irrelevant and even harmful to our purpose.  Therefore, a Skeleton-based Pose Adapter method is proposed to get the aligned pose sequence $\mathbf{Q}$, which decouples the action information from the template pose sequence and embeds the identity information into the aligned $\mathbf{Q}$. The overall logic of our algorithm is presented in Algorithm~\ref{alg:Pose Adapter}.

Formally, the extracted template pose sequence $\mathbf{P} = \{\mathbf{p}_a \}^{M}_{a=1}$, where $\mathbf{P}$ includes the position information of the human keypoints. And the skeletal pose information $\mathbf{q_0}$ estimated from the reference image is denoted as $\mathbf{q_0}$.
Let $\mathbf{C_1} = \{\mathbf{c}_{1i}\}^n_{i=1}$ and $\mathbf{C_2} = \{\mathbf{c}_{2i}\}^n_{i=1}$ be two sets of coordinates representing the key points of $\mathbf{p}_a$ and $\mathbf{q_0}$, where $n$ is the number of points in each set. Let $\mathbf{L} = \{(i_k, j_k)\}^m_{k=1}$ be a sequence of limb connections between points in $\mathbf{C_1}$ and $\mathbf{C_2}$. For each pair of connected points $(i_k, j_k) \in \mathbf{L}$, we calculate the Euclidean distances $d_{1_{k}}$ and $d_{2_{k}}$ in $\mathbf{C_1}$ and $\mathbf{C_2}$ to get the ratio $r_k$:
\begin{align}
r_k = \frac{d_{2_{k}}}{d_{1_{k}}} = \frac{\left\| \mathbf{c}_{1j_k} - \mathbf{c}_{1i_k} \right\|_2}{\left\| \mathbf{c}_{2j_k} - \mathbf{c}_{2i_k} \right\|_2},
\end{align}
so the vector of all ratios is expressed as $\mathbf{r} =  \frac{\mathbf{d_2}}{\mathbf{d_1}} =  \{ r_k \}^m_{k=1}$.

Given two vectors \(\mathbf{c_{1j_k}}\) and \(\mathbf{c_{1i_k}}\) representing the start and end coordinates of a limb segment, we can calculate the length of the limb and the angle between the vectors using the following formula:
\begin{align}
\theta_k = \arctan2\left(  \mathbf{c}_{1j_k} - \mathbf{c}_{1i_k}, { \mathbf{c}_{2j_k} - \mathbf{c}_{2i_k}} \right).
\end{align}

Considering the process of moving a point \(\mathbf{P}\) in a two-dimensional space along a direction determined by an angle \(\theta_k \) by an updated distance $l_k = r_k * d_{1k}$, we first define a vector \(\mathbf{v_k}\) whose magnitude and direction are determined by $l_k$ and $\theta_k$:
\begin{align}
\mathbf{v_k} = l_k \cdot \begin{bmatrix} \cos(\theta_k) \\ \sin(\theta_k)\end{bmatrix}.
\end{align}
Finally, the coordinates of the new point \(\mathbf{c_{2i}}'\) can be obtained by adding the original point \(\mathbf{c_{1i}}\) to the vector \(\mathbf{v_k}\), which can be represented as:
\begin{align}
\mathbf{c_{2i}}' = \mathbf{c_{1i}} + \mathbf{v_k} + \mathbf{\epsilon},
\end{align}
where $\mathbf{\epsilon}$ is the offset of the base coordinate. When set to 0, the position of aligned pose stays the same of the template pose. When set it to the difference between the base coordinates of $\mathbf{C_1}$ and $\mathbf{C_2}$, the position of the aligned pose is consistent with the reference character. Through the above process, we can obtain the aligned pose $\mathbf{C_2}'$ of every frame, to finally construct the aligned pose sequence Q.

\subsection{Kickstart Alignment Strategy}
\label{subsec:Kickstart Alignment}
Inspired by this concept, our approach further enhances the alignment of the input reference image through a similar kickstart alignment technique. We achieve this by employing pose-guided image synthesis models, specifically PCDMs~\cite{shen2023advancing}. 
By doing this, we make a more accurate and natural depiction of the initial frame of generated video. This strategy ensures that the animated character's starting position is  poised to transition seamlessly into the animated sequence, much like a dancer's initial stance before an expressive performance.

The kickstart alignment involves an initial alignment using the first frame of pose sequences to identify key points and skeletal structures from the reference image. This step lays the groundwork for the subsequent pose-guided generation, ensuring that the reference image's pose is conditioned on the first frame of an adjusted pose sequence. This frame serves as a control signal, guiding the reference image to mimic the specific action depicted in the pose. The selection of the initial pose frame is motivated by its role in setting the tone for the entire animation, much like a dancer's initial stance sets the stage for their performance.

Our method's utilization of pose-controlled generation models enables a high degree of control over  pixel-level alignment, ensuring that the animated output is not only consistent with the motion sequence but also preserves the visual integrity of the reference image. This dual emphasis on pose and pixel alignment leads to a more natural and seamless animation.

\section{Experiments}
\label{exp}
\subsection{Experiment Settings}
We implement the experiments based on the existing pose guided video generation Model. For each character animation, we set the reference image with a unified 768×512 resolution. The template videos can be a different resolution. All experiments are performed on a single NVIDIA A100 GPU. Since our method only aligns the input conditions and is training-free, the experiments we conduct are all ablation studies to verify the effectiveness of the proposed method. 

\begin{figure*}[htbp]
	\centering
	\includegraphics[width = 1.0\linewidth]{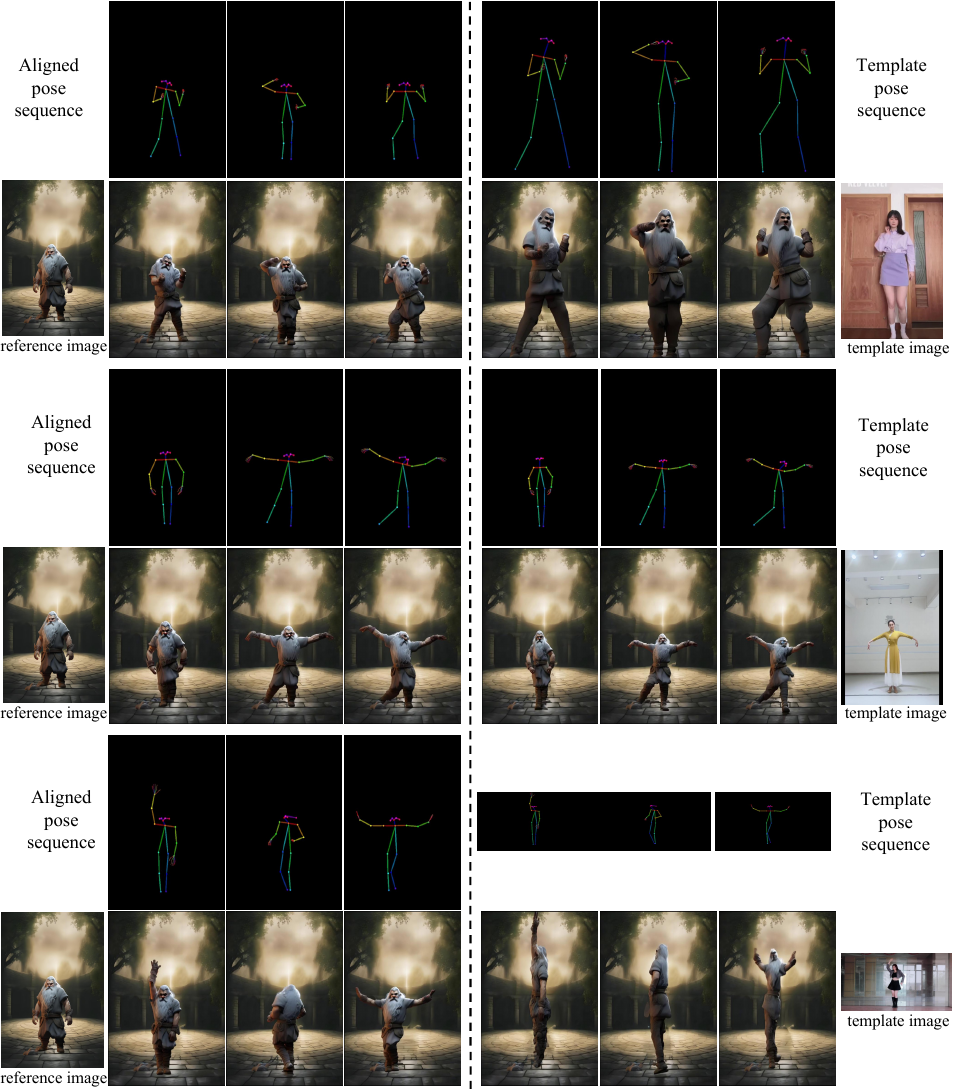}
	\caption{Ablation experiments of Pose Adapter on anime video generation.}
\label{fig:visual_dawlf}
\end{figure*}

\begin{figure*}[htbp]
	\centering
	\includegraphics[width = 1.0\linewidth]{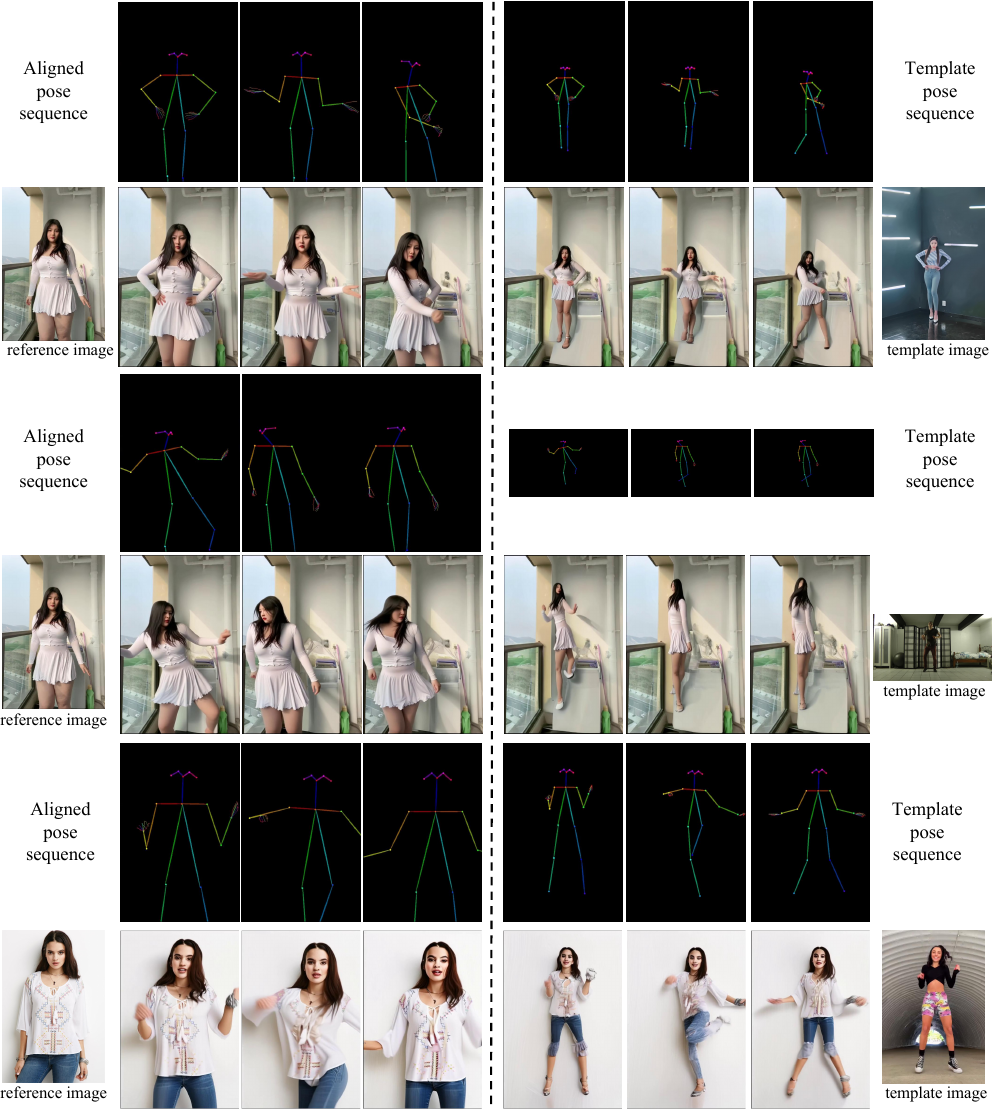}
	\caption{Ablation experiments of Pose Adapter on human video generation.}
\label{fig:visual_human}
\end{figure*}

\begin{figure*}[htbp]
	\centering
	\includegraphics[width = 0.7\linewidth]{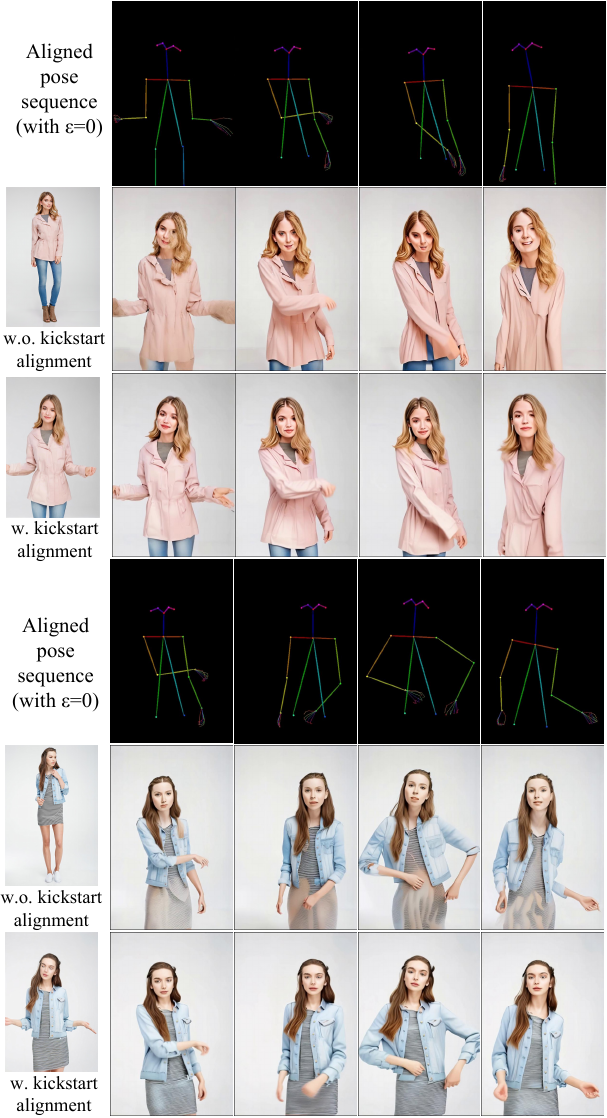}
	\caption{Ablation experiments of Kickstart Alignment on human video generation.}
\label{fig:visual_kickstart}
\end{figure*}

\subsection{Comparison Result}
 To ensure a fair comparison, we employed the same base video generation network architecture, network weights, and test dataset. The following are comparative experiments for the proposed modules.

\textbf{Comparison experiments of Skeleton based Pose Adapter.  } To evaluate the performance of our Skeleton based Pose Adapter, we conducted experiments driving by pose sequences from templates and pose sequences aligned by the Pose Adapter, respectively. The results are displayed in Figure~\ref{fig:visual_dawlf} and Figure~\ref{fig:visual_human}, which correspond to animations of anime characters and humankind in realword, respectively. On the left side, the poses sequence with the Pose Adapter and the generated animation video are presented, on the right side, there are template pose images and the output without the Pose Adapter. It is evident that when the template poses are not aligned with the input, the generated results are quite poor. 

In Figure~\ref{fig:visual_dawlf}, as shown in the first and second sets. The base model is unable to address the discrepancy in body shape between the template and the reference image, resulting in generated frames altering the original identity characteristics of the input image. For instance, a dwarf loses its distinctive stocky physique and instead assumes a body shape similar to that of a human. And the third set, when the frame size is inconsistent, the pose image is squeezed and deformed, losing its control ability. Similar situation also appears in Figure~\ref{fig:visual_human}. 
When the character's position in the frame is misaligned, the person and the background will become intertwined. When the frame size is inconsistent (the same as the third set of Figure~\ref{fig:visual_dawlf}), the pose image is deformed, and the results collapses entirely. When there is a mismatch between full-body poses and half-body reference, sometimes the animation result will be difficult to accept. 

\textbf{Comparison experiments of Kickstart Alignment.  }
In this section, we further incorporated Kickstart Alignment, which aligns the reference image with the gesture of the first pose of the template sequence.  Figure~\ref{fig:visual_kickstart} presents the results of two sets of Kickstart Alignment on human video generation, where the first row of each set is the aligned pose sequence. In this part, we set $\epsilon$ to 0 to prevent misalignment effects between half-body and full-body poses. 
The first set of results clearly illustrates that the absence of Kickstart Alignment results in the collapse of facial features and hair in the generated images. Moreover, the subsequent set of results indicates that the lack of Kickstart Alignment may also give rise to undesirable texture alterations. Generative models are tasked with the formidable challenge of extracting essential human body information from unaligned reference images and subsequently incorporating this information into the generation process. However, current models find this task to be overly demanding. Our approach, which incorporates alignment at the outset, effectively alleviates this challenge, resulting in substantial improvements in the quality of generated results.

\section{Conclusion}
\label{con}
In this paper, we present a novel training-free augmentation strategy for generating pose-guided personalized videos. To tackle the misalignment between original videos and reference characters, we introduce two critical algorithms: Skeleton-based Pose Adapter and Kickstart Alignment strategy. The visualization results indicate that our method exhibits a significant improvement on image fidelity to the source image while preserving intricate fine-grained appearance details. Our approach relies solely on input control conditions and does not require extra training, enabling straightforward integration into a wide variety of pose-guided video generation models. Moreover, our method involves only basic linear matrix operations and the creation of single-frame images, making it highly efficient.

\bibliography{icml2024}
\bibliographystyle{icml2024}

\end{document}